\documentclass[letterpaper, 10pt, conference]{IEEEtran}
\IEEEoverridecommandlockouts
\usepackage{cite}
\usepackage{amsmath,amssymb,amsfonts}
\usepackage{algorithmic}
\usepackage{graphicx}
\usepackage{textcomp}
\usepackage{xcolor}
\usepackage{booktabs}
\usepackage{tabularx}

\def\BibTeX{{\rm B\kern-.05em{\sc i\kern-.025em b}\kern-.08em
    T\kern-.1667em\lower.7ex\hbox{E}\kern-.125emX}}
\begin{document}

\title{IntTrajSim: Trajectory Prediction for Simulating Multi-Vehicle driving at Signalized Intersections\\
}

\author{\IEEEauthorblockN{1\textsuperscript{st} Yash Ranjan}
\IEEEauthorblockA{\textit{Department of CISE} \\
\textit{University of Florida}\\
yashranjan@ufl.edu}
\and
\IEEEauthorblockN{2\textsuperscript{nd} Rahul Sengupta}
\IEEEauthorblockA{\textit{Department of CISE} \\
\textit{University of Florida}\\
rahulseng@ufl.edu}
\and 
\IEEEauthorblockN{3\textsuperscript{rd} Anand Rangarajan}
\IEEEauthorblockA{\textit{Department of CISE} \\
\textit{University of Florida}\\
anand@cise.ufl.edu}
\and
\IEEEauthorblockN{4\textsuperscript{th} Sanjay Ranka}
\IEEEauthorblockA{\textit{Department of CISE} \\
\textit{University of Florida}\\
ranka@cise.ufl.edu}
}

\maketitle

\begin{abstract}
Traffic simulators are widely used to study the operational efficiency of road infrastructure, but their rule-based approach limits their ability to mimic real-world driving behavior. Traffic intersections are critical components of the road infrastructure, both in terms of safety risk (nearly 28\% of fatal crashes and 58\% of nonfatal crashes happen at intersections) as well as the operational efficiency of a road corridor. This raises an important question: can we create a data-driven simulator that can mimic the macro- and micro-statistics of the driving behavior at a traffic intersection? Deep Generative Modeling-based trajectory prediction models provide a good starting point to model the complex dynamics of vehicles at an intersection. But they are not tested in a "live" micro-simulation scenario and are not evaluated on traffic engineering-related metrics. In this study, we propose traffic engineering-related metrics to evaluate generative trajectory prediction models and provide a simulation-in-the-loop pipeline to do so. We also provide a multi-headed self-attention-based trajectory prediction model that incorporates the signal information, which outperforms our previous models on the evaluation metrics.
\end{abstract}

\begin{IEEEkeywords}
Traffic Intersection, Simulation, SUMO, Signal, ITS, Machine Learning, Deep Generative Modeling
\end{IEEEkeywords}

\section{Introduction}
The traffic intersection plays a critical role in the transportation infrastructure in terms of safety. As it is the place where pedestrians interact with vehicles, there is a higher chance of accidents. Overall, it is critical for the operational efficiency of the road infrastructure, as the signal timing at an intersection influences the travel time across a corridor. Traffic simulators such as SUMO\cite{SUMO2018} and VISSIM\cite{Fellendorf2010} have been used successfully to study the operational efficiency of transportation infrastructure. But the vehicle dynamics controlled in these simulators is based on handcrafted logic that makes the vehicles and pedestrians follow rules i.e., following the signal, avoiding collisions, yielding at intersections, maintaining safe distances, and adhering to speed limit \cite{chen2023datadriventrafficsimulationcomprehensive}. These simulators are unable to capture the latent interactions between the vehicle and the dynamic and static actors around it, and how it affects the dynamics of the vehicle. Trajectory prediction has been used in the field of Autonomous Driving to provide a plausible set of future trajectories for vehicles surrounding the ego vehicle so that the ego vehicle can plan its short-term control dynamics properly. But trajectory prediction models have not been used to simulate multi-vehicle driving at an intersection before. That is because trajectory prediction models are evaluated on trajectory reconstruction errors, which do not consider the traffic-engineering related concerns like red-light violation, collisions, mid-intersection stoppages, etc. Moreover, they do not take the signal timing plan as an input to predict future trajectories.

In this paper, we present a framework to train a trajectory prediction model while incorporating actors that are specific to a traffic intersection, and run inference on the model in a closed loop to simulate multi-vehicle driving. We propose a Conditional Variational Auto-Encoder-based\cite{kingma2022autoencodingvariationalbayes}  trajectory prediction model that uses a vector-based representation of the scene as well as a multi-headed attention \cite{vaswani2023attentionneed} encoder to incorporate the influence of all the agents surrounding the vehicle, including the traffic signal. We also propose a set of traffic-related metrics on top of the ones we published at \cite{Ranjan2025PAKDD} that allows us to evaluate the trajectories generated by the models using the simulation-in-the-loop method.

\section{Related Work}
We are using Trajectory Prediction framework to create a micro-simulator for a traffic intersection scene. Trajectory prediction is a very well-studied topic. Physics-based models\cite{8186191} used equations from physics and kinematics to find the future state of objects. The Kalman Filter approach\cite{9234702} has been used for trajectory prediction and tracking tasks. These traditional approaches are computationally efficient and also provide explainable results but do not work well in complicated scenarios.

Deep Learning based approaches have found great success in recent years, especially due to the abundance of data collected via video cameras, LiDAR, and Loop Detectors. LSTM-based approaches\cite{graves2014generatingsequencesrecurrentneural}\cite{8569595} were able to capture the temporal relationship of the vehicle history to predict the future states. One of the earlier pioneering works to capture the dynamic correlation between objects in the scene to output socially-aware trajectories was Social LSTM\cite{7780479}.

With the advent of Generative modeling, approaches using GANs and VAEs, which output a probability distribution over future states, became popular. Trajectron\cite{ivanovic2019} and Trajectron++\cite{salzmann2021} are two important works that represented traffic scene as a spatial-temporal graph to produce dynamically feasible trajectories. Target-driven trajectory prediction\cite{zhao2020tnttargetdriventrajectoryprediction} was also an important work as it led to intent conditioned trajectory prediction frameworks.

In recent works, transformer\cite{vaswani2023attentionneed} based architecture for trajectory prediction has been widely used due to its scalability, ease of training, and generalization ability. MTR\cite{shi2023motiontransformerglobalintention} and DenseTNT\cite{gu2021densetntendtoendtrajectoryprediction} are SOTA models that use transformer architecture to predict future trajectories of objects.

\begin{figure}[!htbp]
\centering
\includegraphics[width=0.45\textwidth]{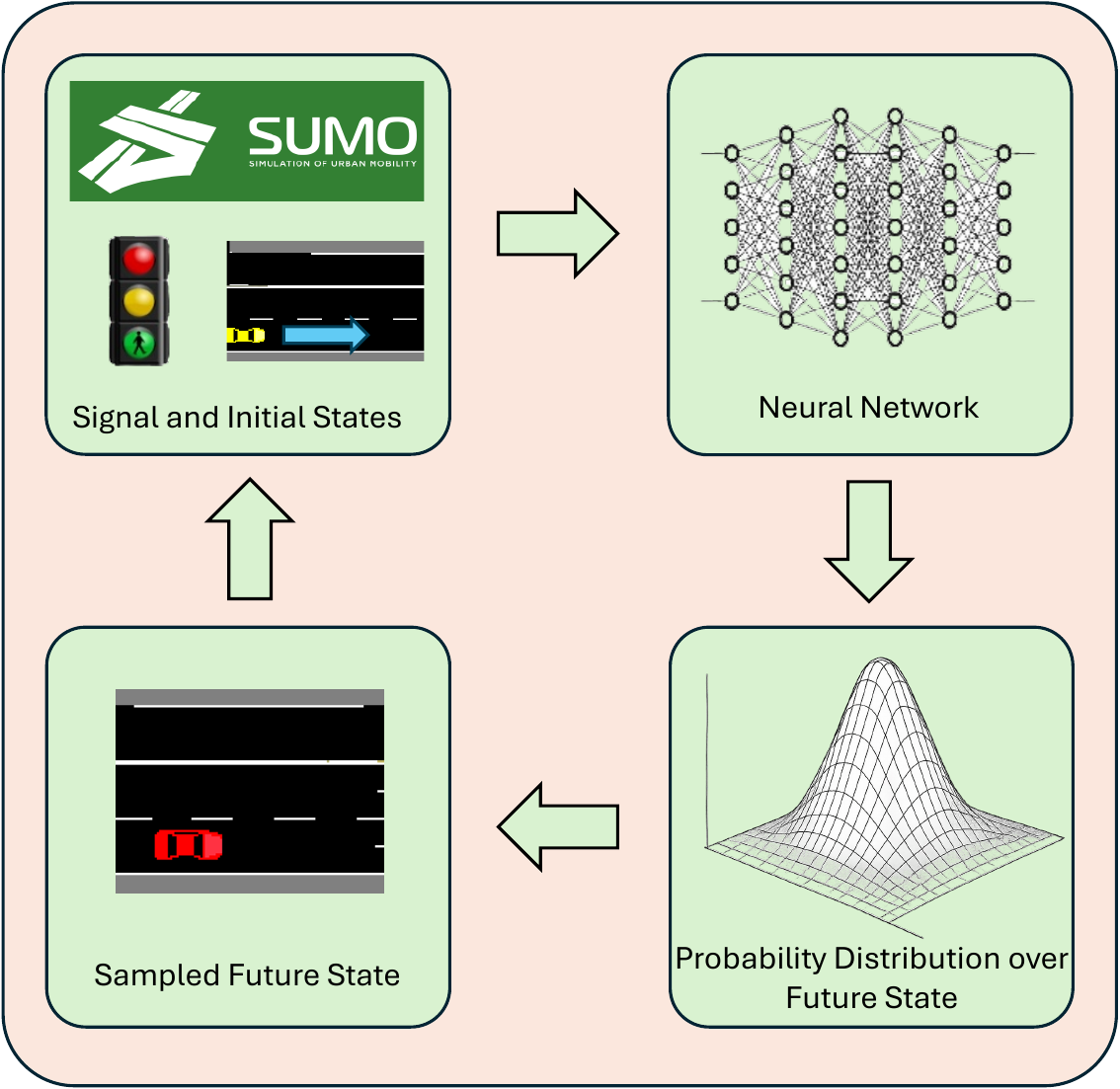}
\caption{Simulation-in-the-loop framework to generate trajectories by predicting distribution over future states conditioned on surrounding actors.}
\label{fig:data_flow}
\vspace{-3mm}
\end{figure}

\section{Problem Statement}
We want to develop a simulator for traffic intersections that takes into account the multiple actors (static as well as dynamic) present at the intersection and is able to generate trajectories that mimic the actual driving behavior based on ground truth trajectories. In an ego-centric view of the world, the driver's actions are influenced by:
\subsubsection{Neighboring Dynamic Agents} Dynamic agents are the multiple neighboring vehicles surrounding the driver. In a traditional car following model, the vehicles only react to the vehicle right in front of it in their lane so as to maintain a safe distance from it. But in the real world, a driver needs to observe every vehicle in it's visible range so as to preempt their actions and to accordingly control its own state.
\subsubsection{Signal Control Variable} The signal state at a traffic intersection is a very important factor that influences a driver's actions. In the car-following model, the vehicle only cares about the signal state when there is no vehicle in front of it. If there is a vehicle in front of it, then it only reacts to that vehicle and not the signal. In the real world, a driver needs to pay attention to the signal irrespective of their neighbor, as breaking the signal can cause dangerous situations like accidents.
\subsubsection{Intersection Infrastructure} The lane markings, stop lines and road boundaries provide a navigable region for the driver to drive on. In traditional simulators, vehicles are supposed to follow the lane as a rule. But in real-world scenarios, if the driver is not paying attention to these markings, it can lead to dangerous situations like going off the road or unintentionally going off the lane to the next lane.
\subsubsection{Own Previous State} The action taken by a driver is dependent on the previous state of the vehicle. Although every human being is unique and can take different actions given the same situation, we can still model it using a probability distribution that can capture the most likely action taken for the situation. In physics-based simulators, these distributions are governed by hard-coded parameters and are not learnt based on data. Moreover, we cannot model the driver's action based on time of the day or weather conditions, which we can do using a data-driven simulator as the data will capture these latent behaviors.
\subsubsection{Attention} In the real world, a driver observes all these factors and subconsciously assigns importance (dynamically) to them and takes actions based on what it finds most important. A physics-based simulator decides what a vehicle should focus on based on a physics equation. This causes it to not be able to capture situations where the driver is drunk or distracted on the phone and is not able to pay attention to one or some of these factors, leading to a mistake that can even be fatal.

From the above problem statement, it becomes very clear that it is important to study the application of data-driven modeling approaches to create a simulator for traffic intersections that simulates the multi-agent driving behavior based on ground truth data. Next, we propose a novel deep learning-based architecture that incorporates these factors to generate set of trajectories which are feasible from a traffic intersection perspective.

\section{Attention-Based Trajectory Prediction for Signalized Intersection}

We are using a Conditional Variational Auto Encoder\cite{kingma2022autoencodingvariationalbayes} architecture to learn a distribution over future states of vehicle conditioned on its previous states, previous states of the neighboring vehicles as well as static objects around it, such as the lane marking and traffic signal. We utilize a multi-headed self-attention \cite{vaswani2023attentionneed} mechanism to create an embedding of the state of the vehicle that captures the influence that the surrounding vehicles, as well as static objects, have on the state of the vehicle. We model the probability distribution over future states using Gaussian Mixture Model\cite{viroli2017deepgaussianmixturemodels} whose parameters are learned from the data.

\subsection{Encoder}
\subsubsection{Actor Vehicle History Encoder}
A vehicle's previous states are important for generating future trajectories, as the motion of a vehicle must follow dynamic constraints to ensure smoothness. To capture the temporal correlation between the previous states of the vehicle, we employ the LSTM architecture where we pass the vehicle's rectangular coordinates, the x and y components of its velocity, and the x and y components of its acceleration.  The coordinates are normalized to be centered at the current location of the vehicle.
\begin{equation}
    h^t_i = LSTM \left( h^{t-1}_i, {x}^t_i, W_{hist} \right),
\label{eq:hist_enc}
\end{equation}
where ${x}^{t}_i$ is the vehicle state and $h^t_i$ is the vehicle history embedding.

While training, we encode the future ground truth trajectory. We use a bi-directional LSTM to get an embedding $h^t_{i, fut}$ for the future trajectory. We use this embedding to model the latent distribution.

\begin{figure*}[!htbp]
\centering
\includegraphics[width=0.90\textwidth]{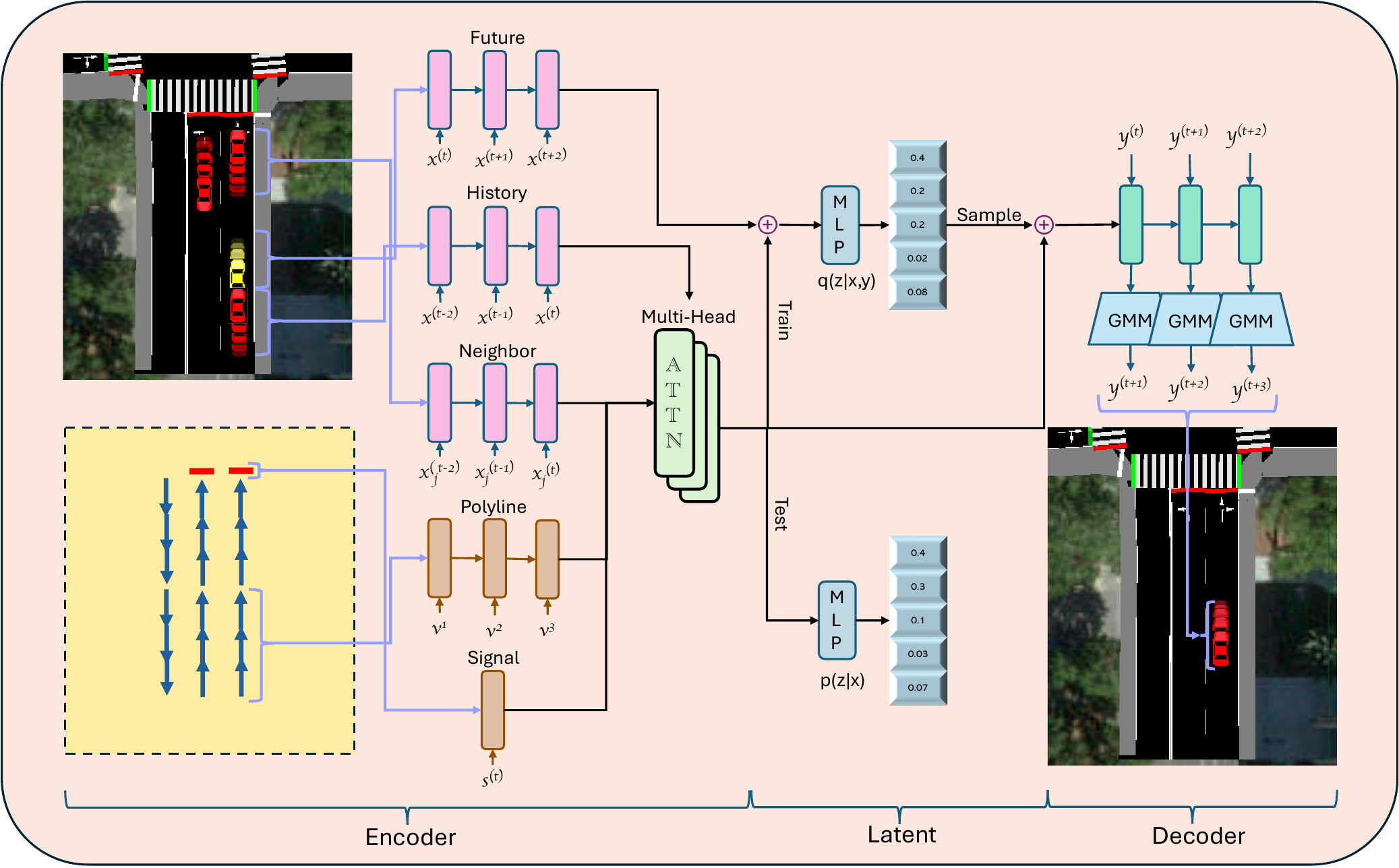}
\caption{CVAE based architecture for trajectory prediction at an Intersection that incorporates the states of actors using Multi-Headed Attention and outputs a Mixture of Gaussian distribution over the future states.}
\label{fig:architecture}
\vspace{-3mm}
\end{figure*}

\subsubsection{Neighbor Vehicle History Encoder}
The actor vehicle will observe the vehicles in its surroundings and will control its own state based on the previous states of the neighbors. Hence, for each neighbor $k$ for the actor vehicle $i$, we create an encoded representation of the previous states of the neighboring vehicle. We employ the LSTM architecture and pass the neighboring vehicle's coordinates, velocity, and acceleration. But all state components of the neighbors are calculated relative to the actor vehicle, as the actor vehicle's future trajectories are influenced by the state of the neighboring vehicle that it observes and not the actual ground states.
\begin{equation}
h^t_{i,k} = LSTM \left( h^{t-1}_{i,k}, n^t_{i,k}; W_{nbr} \right), \quad\forall k \in N_i 
\label{edge_encoder}
\end{equation}

where $n^t_{i,k}$ is the state of the neighbor $k\in N_i$ observed by the vehicle $i$ and $h^t_{i,k}$ is the embedding of the neighbor vehicle history. 

\subsubsection{Lane Center Line Encoder}
The lane center line provides an estimate of the driving region to the driver of a vehicle. We need to encode the lane center lines so that the model can utilize the information to learn a better distribution over the future trajectories\cite{gao2020vectornetencodinghdmaps}. Earlier models like \cite{salzmann2021} were encoding the road map using rasterized processing, where maps are encoded using Convolution Neural Networks\cite{oshea2015introductionconvolutionalneuralnetworks}. VectorNet \cite{gao2020vectornetencodinghdmaps} introduced an efficient polyline scene representation to capture high-definition (HD) map features. We are using a similar representation for the lane center lines where the lanes are divided into polylines and each polyline comprises of collection of connected vectors. A vector is represented by the starting coordinates (centered around the actor vehicle), the length of the vector, and the direction of the vector, represented by the sin and cos values of the angle. To keep the polylines consistent, each polyline consists of three connected vectors, and the length of each vector is 4 meters. Each polyline is passed through an LSTM to obtain an embedded representation.
\begin{equation}
    p^l_{i, j} = LSTM \left( p^{l-1}_{i, j}, v^{l}_{i, j}; W_{lane}\right),\quad \forall j \in P_i
\label{polyline_encoder}
\end{equation}
where $v^{l}_{i, j}$ is the vector $l$ belonging to the polyline $j$ as seen by vehicle $i$ and $p^l_{i, j}$ is the embedding for polyline $j$.

\subsubsection{Traffic Signal Encoder}
The traffic signal plays an important role in controlling the driving behavior of vehicles at an intersection. Traditional simulators do not allow us to understand the conditions that causes the driver to follow or to not follow the traffic signal. To learn the influence of traffic signal on the vehicle, we encode the signal for the lane in which the vehicle is in. We pass the coordinates of the end of the stop line for the lane and the one-hot encoded traffic signal through a feed-forward neural network.
\begin{equation}
    s^n_{i} = MLP \left( s^n_{i, 0}, s^n_{i, 1}, tl^t_i \right),
\label{signal_encoder}
\end{equation}
where $s^n_{i, 0}$ and $s^n_{i, 1}$ are the stop line coordinates and $tl^t_i$ is the traffic light for that lane at time $t$.

\subsubsection{Multi-Headed Self-Attention}
A driver should ideally focus on all the above factors while controlling its state. But many a time, a driver can miss out on some of these factors. To capture the correlation that the dynamic factors, such as neighboring vehicles and static factors like lane center-line and signal, have with the actor vehicle, and how it influences it, we are using multi-head self-attention \cite{vaswani2023attentionneed}.
\begin{equation}
\text{Attention}(Q, K, V) = \text{softmax}\left(\frac{QK^T}{\sqrt{d_k}}\right)V,
\end{equation}
where $Q, K$ and $V$ are Query, Key and Value respectively and $d_k$ is the embedding size which is 128. The query vector is a linear projection of the encoded actor vehicle history $h^t_i$, and the key and value vectors are the linear projections of the neighbor $h^t_{1,k}$ $\forall k\in N_i$, lane center-line $p^l_{i,j}$ $\forall j\in P_i$, and traffic signal encoded vector $s^n_i$ concatenated together.
\begin{align}
    Q &= \text{Linear}(h^t_i) \\
    K &= \text{Linear}(\text{Concat}(h^t_{i,1..k},\ p^l_{i,1..j},\ s^n_i)) \\
    V &= \text{Linear}(\text{Concat}(h^t_{i,1..k},\ p^l_{i,1..j},\ s^n_i))
\end{align}

In multi-head attention, multiple attention heads calculate the correlation between the query and key vectors in parallel and then concatenate the output of all the attention heads. We use 4 attention heads to train the models.
\begin{align}
    \text{head}_i = \text{Attention}(Q W_i^Q, K W_i^K, V W_i^V) \\
    \text{MultiHead}(Q, K, V) = \text{Concat}(\text{head}_1, \dots, \text{head}_h) W^O
\end{align}

As per \cite{vaswani2023attentionneed}, we concatenate the output of the attention heads and then sum the final embedding to the query vector (actor vehicle history embedding) to get the final embedding. This way, the embedding represents the state of the vehicle at that time as well as captures the influence of the dynamic and static agents around it.

\begin{equation}
    h^t_{i,enc} = h^t_i + \text{MultiHead}(Q, K, V)
\end{equation}

\subsection{Latent Variable Modeling}
Using the embedding from the encoder $h^t_{i,enc}$ we parametrize the prior distribution $p_\psi(z|x)$ and we approximate the true posterior by parameterizing $q_\phi(z|x,y)$ using the $h^t_{i,enc}$ and $h^t_{i,fut}$. We keep the latent variable z discrete so as to model different modes for predicting future trajectories y given the input condition x. The conditional probability distribution \( p(y \mid x) \) over all z will be:
\begin{equation}
    p(y|x) = \sum_{z \in Z} p_\theta( y| x, z) p_\psi (z| x)),
\end{equation}

Instead of producing a deterministic trajectory prediction, the model is able to predict a distribution of plausible future trajectories, accounting for the variability of vehicle behavior. For example, when drivers see a red light, it can either start decelerating from a distance and come to a stop earlier, or it can decelerate much later and come to a stop near the stop line, or in some cases, not decelerate at all. Under different situations, the probability of each behavior varies. We set number of modes $|Z|$ to be 25 in our experiments.

\subsection{Decoder}
We use a decoder network to predict a distribution over the future state of the vehicle based on the final embedding $h^t_{i,enc}$ we get from the Encoder and the sampled latent variable $z$. $z$ is sampled from the parametrized distribution $q_\phi(z|x,y)$ while training and from $p_\psi(z|x)$ while testing. 
\begin{equation}
z \sim
\begin{cases}
q_\phi(z \mid x_i, y_i), & \text{for training} \\
p_\psi(z \mid x_i), & \text{for testing}
\end{cases}
\end{equation}
We concatenate the $z$ and $h^t_{i,enc}$ and pass it through a GRU neural network with a hidden dimension of 128 to get the decoded embedding. The decoded vector is then passed through a mixture density network \cite{370fbeadb5584ba9ab2938431fc4f140} to get the mean and covariance of the Gaussian for each $z$ component for the next time-step. We take the mean and covariance and use it as parameters for the Gaussian Mixture Model to model the distribution $p_\theta ({y} | {x}, z)$ over the next 20 timesteps. 
\begin{equation}
\hat{y}_i^t \sim GMM\left( GRU\left( [\hat{y}_i^{t-1}, z, h_{i,enc}^t] ; W_{\theta} \right) \right)
\end{equation}
For model training, we employ the CVAE's objective function, as described in \cite{zhao2018infovaeinformationmaximizingvariational} with the Adam optimizer:

\begin{equation}
\begin{aligned}
    \max_{\phi, \theta, \psi} \sum_{i=1}^{N} \mathbb{E}_{z \sim q_\phi(z \mid x_i, y_i)}
    \left[ \log p_\theta(y_i \mid x_i, z) \right] \\
    {} {}-\beta D_{KL}\left(q_\phi(z \mid x_i, y_i) \parallel p_\psi(z \mid x_i)\right)
\end{aligned}
\label{eq:loss_func}
\end{equation}

During inference, we sample the probability distribution $y \sim p_\theta ({y} | {x}, z)$ to obtain the future trajectories. The sampling can be performed by either finding the mixture weights of the GMM from the model \( p_\psi(z\mid x)\)  and then sampling the most likely output of the distribution $y\sim p_\theta ({y} | {x}, z)$ or sampling the individual Gaussian to obtain the trajectories for the different modes.

\begin{figure}[!htbp]
\centering
\includegraphics[width=0.45\textwidth]{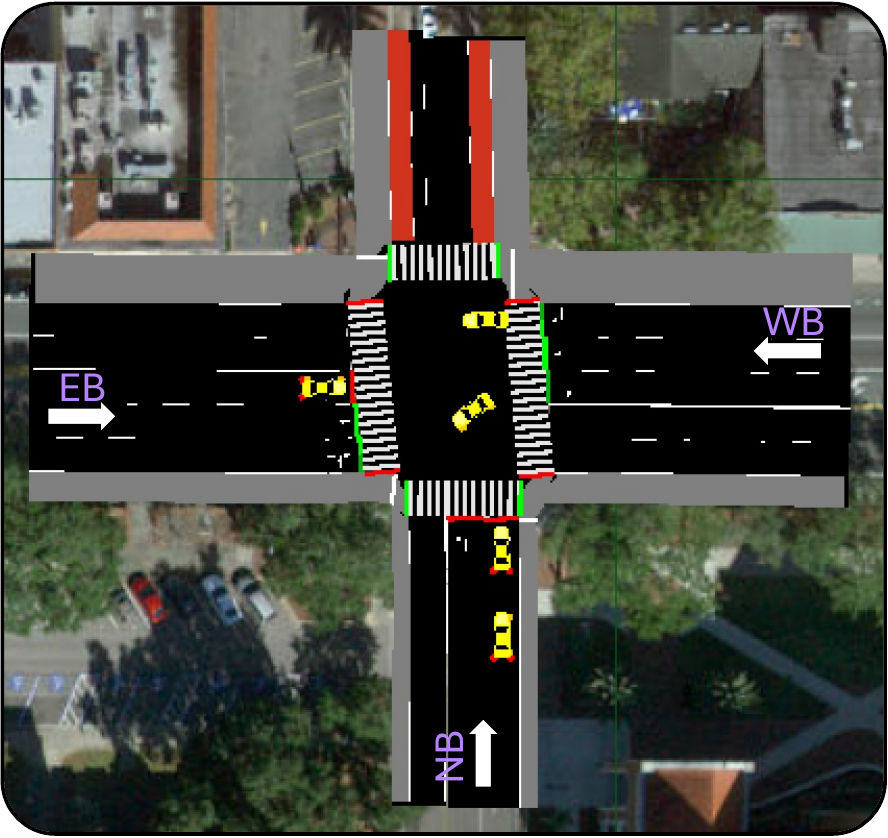}
\caption{Digital Map Created for the intersection at West Univ Ave @ NW 17th Street in Gainesville, Florida}
\label{fig:map}
\vspace{-3mm}
\end{figure}
\subsection{Training}
We have created a digital map for the traffic intersection at West Univ Ave @ NW 17th Street in Gainesville, Florida, in SUMO (Figure \ref{fig:map}). We generate vehicle trajectories using the randomTrips.py file in SUMO, keeping period as 0.5, fringe-factor 10, and binomial 2. We give traffic signal timing similar to what we observe at the intersection so as to keep the traffic flow as close to the real world as possible. To train the model, we generate 6 hours (21600 seconds) of data simulated at a frequency of 1/10th of a second. The total number of exemplars (number of vehicles at each timestamp * number of timestamps) is 1,241,600. The output embedding size of each encoder is 128, which is passed to the multi-headed self-attention with 4 attention heads.

Since we generate the data from SUMO, which uses Krauss Model \cite{etde_627062} to simulate the vehicle motion, we put in some constraints on the information that each vehicle has at each timestamp so as to reduce the amount of noise in the predicted output. Every vehicle only considers the two nearest vehicle that is within its attention radius and is in front of it in the lane. Every vehicle gets the next two polyline information that are going to come in its route. This helps the model learn a better distribution over the future trajectories and helps reduce outliers over long-term trajectory unrolling. But the architecture of the model can be scaled to train on data where these constraints are not required, such as real-world trajectory data at an intersection.

\begin{table*}[ht]
\centering
\small
\caption{Evaluation Metric Results from Current Model}
\begin{tabularx}{\textwidth}{l *{7}{>{\centering\arraybackslash}X}}
\toprule
Trajectory Cluster & Total Count & Red Light Violation & Mid-Intersection Stoppage & Pre-Stopbar Stoppage & Unsafe Deceleration & Reversing & TTC Events \\
\midrule
L on EBL & 102 & 0 (0.0\%) & 0 (0.0\%) & 0 (0.0\%) & 4 (3.9\%) & 16 (15.7\%) & 98 \\
L on NBL & 133 & 0 (0.0\%) & 1 (0.8\%) & 0 (0.0\%) & 0 (0.0\%) & 0 (0.0\%) & 27 \\
L on WBL & 127 & 39 (30.7\%) & 0 (0.0\%) & 0 (0.0\%) & 0 (0.0\%) & 0 (0.0\%) & 36 \\
R on EBTR & 99 & 0 (0.0\%) & 0 (0.0\%) & 0 (0.0\%) & 4 (4.0\%) & 4 (4.0\%) & 108 \\
R on NBTR & 115 & 0 (0.0\%) & 0 (0.0\%) & 0 (0.0\%) & 0 (0.0\%) & 0 (0.0\%) & 0 \\
R on WBTR & 106 & 69 (65.1\%) & 0 (0.0\%) & 0 (0.0\%) & 0 (0.0\%) & 0 (0.0\%) & 13 \\
T on EBT & 51 & 0 (0.0\%) & 0 (0.0\%) & 0 (0.0\%) & 2 (3.9\%) & 0 (0.0\%) & 19 \\
T on EBTR & 67 & 1 (1.5\%) & 0 (0.0\%) & 0 (0.0\%) & 1 (1.5\%) & 6 (9.0\%) & 29 \\
T on NBTR & 111 & 0 (0.0\%) & 0 (0.0\%) & 0 (0.0\%) & 0 (0.0\%) & 0 (0.0\%) & 38 \\
T on WBT & 22 & 11 (50.0\%) & 0 (0.0\%) & 0 (0.0\%) & 0 (0.0\%) & 0 (0.0\%) & 4 \\
T on WBTR & 93 & 49 (52.7\%) & 0 (0.0\%) & 0 (0.0\%) & 0 (0.0\%) & 0 (0.0\%) & 6 \\
\midrule
Total & 1026 & 169 (16.5\%) & 1 (0.1\%) & 0 (0.0\%) & 11 (1.1\%) & 26 (2.5\%) & 378 \\
\bottomrule
\end{tabularx}
\label{tab:result_new_model}
\end{table*}

\begin{table*}[ht]
\centering
\small
\caption{Evaluation Metric Results on inference data from Earlier Model}
\begin{tabularx}{\textwidth}{l *{7}{>{\centering\arraybackslash}X}}
\toprule
Trajectory Cluster & Total Count & Red Light Violation & Mid-Intersection Stoppage & Pre-Stopbar Stoppage & Unsafe Deceleration & Reversing & TTC Events \\
\midrule
L on EBL & 102 & 0 (0.0\%) & 0 (0.0\%) & 69 (67.6\%) & 0 (0.0\%) & 4 (3.9\%) & 103 \\
L on NBL & 133 & 0 (0.0\%) & 1 (0.8\%) & 47 (35.3\%) & 54 (40.6\%) & 20 (15.0\%) & 113 \\
L on WBL & 127 & 25 (19.7\%) & 0 (0.0\%) & 1 (0.8\%) & 0 (0.0\%) & 0 (0.0\%) & 64 \\
R on EBTR & 99 & 0 (0.0\%) & 0 (0.0\%) & 0 (0.0\%) & 0 (0.0\%) & 3 (3.0\%) & 114 \\
R on NBTR & 115 & 0 (0.0\%) & 0 (0.0\%) & 20 (17.4\%) & 29 (25.2\%) & 25 (21.7\%) & 22 \\
R on WBTR & 106 & 69 (65.1\%) & 0 (0.0\%) & 3 (2.8\%) & 1 (0.9\%) & 3 (2.8\%) & 37 \\
T on EBT & 57 & 0 (0.0\%) & 0 (0.0\%) & 35 (61.4\%) & 0 (0.0\%) & 3 (5.3\%) & 24 \\
T on EBTR & 61 & 0 (0.0\%) & 0 (0.0\%) & 7 (11.5\%) & 0 (0.0\%) & 5 (8.2\%) & 65 \\
T on NBTR & 111 & 0 (0.0\%) & 0 (0.0\%) & 15 (13.5\%) & 21 (18.9\%) & 19 (17.1\%) & 40 \\
T on WBT & 36 & 13 (36.1\%) & 0 (0.0\%) & 0 (0.0\%) & 0 (0.0\%) & 1 (2.8\%) & 7 \\
T on WBTR & 79 & 45 (57.0\%) & 0 (0.0\%) & 0 (0.0\%) & 0 (0.0\%) & 0 (0.0\%) & 15 \\
\midrule
Total & 1026 & 152 (14.8\%) & 1 (0.1\%) & 197 (19.2\%) & 105 (10.2\%) & 83 (8.1\%) & 604 \\
\bottomrule
\end{tabularx}

\label{tab:result_old_model}
\end{table*}

\begin{table*}[ht]
\centering
\small
\caption{Evaluation Metric Results on raw ground truth data from SUMO}
\begin{tabularx}{\textwidth}{l *{7}{>{\centering\arraybackslash}X}}
\toprule
Trajectory Cluster & Total Count & Red Light Violation & Mid-Intersection Stoppage & Pre-Stopbar Stoppage & Unsafe Deceleration & Reversing & TTC Events \\
\midrule
L on EBL & 102 & 0 (0.0\%) & 0 (0.0\%) & 0 (0.0\%) & 0 (0.0\%) & 0 (0.0\%) & 0 \\
L on NBL & 133 & 0 (0.0\%) & 0 (0.0\%) & 0 (0.0\%) & 0 (0.0\%) & 0 (0.0\%) & 0 \\
L on WBL & 127 & 0 (0.0\%) & 44 (34.6\%) & 0 (0.0\%) & 0 (0.0\%) & 0 (0.0\%) & 16 \\
R on EBTR & 99 & 0 (0.0\%) & 0 (0.0\%) & 0 (0.0\%) & 0 (0.0\%) & 0 (0.0\%) & 0 \\
R on NBTR & 115 & 0 (0.0\%) & 0 (0.0\%) & 0 (0.0\%) & 0 (0.0\%) & 0 (0.0\%) & 0 \\
R on WBTR & 106 & 0 (0.0\%) & 0 (0.0\%) & 0 (0.0\%) & 0 (0.0\%) & 0 (0.0\%) & 0 \\
T on EBT & 51 & 0 (0.0\%) & 0 (0.0\%) & 0 (0.0\%) & 0 (0.0\%) & 0 (0.0\%) & 0 \\
T on EBTR & 67 & 0 (0.0\%) & 0 (0.0\%) & 0 (0.0\%) & 0 (0.0\%) & 0 (0.0\%) & 16 \\
T on NBTR & 111 & 0 (0.0\%) & 0 (0.0\%) & 0 (0.0\%) & 0 (0.0\%) & 0 (0.0\%) & 0 \\
T on WBT & 54 & 0 (0.0\%) & 0 (0.0\%) & 0 (0.0\%) & 0 (0.0\%) & 0 (0.0\%) & 0 \\
T on WBTR & 61 & 0 (0.0\%) & 0 (0.0\%) & 0 (0.0\%) & 0 (0.0\%) & 0 (0.0\%) & 0 \\
\midrule
Total & 1026 & 0 (0.0\%) & 44 (4.3\%) & 0 (0.0\%) & 0 (0.0\%) & 0 (0.0\%) & 32 \\
\bottomrule
\end{tabularx}

\label{tab:result_sumo}
\end{table*}

\section{Experiments and Results}

\subsection{Metrics}
In our previous work \cite{Ranjan2025PAKDD}, we have described the steps to calculate the metrics:

\begin{enumerate}
    \item Red Light Violation: The metric tracks the number of times a vehicle breaks the red light signal.
    \item Mid-Intersection Stoppage: This metric tracks the number of times the vehicle stops in the middle of the intersection.
    \item Pre-Stopbar Stoppage: This metric tracks if the vehicle does not respond properly to signal transition from Red to Green.
    \item TTC Events: This metric tracks the number of times the minimum TTC value between two vehicles is below a threshold (1 second).
\end{enumerate}

We have added new metrics to the list that capture relevant intersection-related scenarios.

\subsubsection{Unsafe Deceleration}
Sudden braking at an intersection is a dangerous event, as it can cause accidents. We want to check if, while unrolling trajectories, we are seeing sudden braking behavior. We have characterized the braking as per deceleration thresholds\cite{vehits23}:

\begin{enumerate}
    \item -0.35g to -0.47g as MILD Braking
    \item -0.47g to -0.62g as HARD Braking
    \item  Beyond –0.62g as EXTREME Braking
\end{enumerate}

And we count the number of times the deceleration is in the range of either Hard braking or Extreme Braking. We are using acceleration due to gravity "g" (9.8$m/s^2$) as the unit of measurement.

\subsubsection{Reversing}
While running the models in inference mode in the simulation-in-loop, we observed that vehicles would start moving in the backward direction. This would usually happen when the model is not able to generalize for the input states. To measure this behavior, we check if the vehicle is moving backward in the lane for more than 10 continuous frames. 


\subsection{Experiments}
For evaluating the model's performance on these metrics, we have created a pipeline where we use the SUMO simulator to control the rate at which the vehicles enter the intersection and the signal state at each timestep. We then allow SUMO to simulate the initial states (starting 20 timesteps in our experiment) for each vehicle and give that as input to the model to start generating future trajectories. We take the generated coordinate for the next timestep and concatenate that to the initial state of the vehicle. We again take that as an input for the model and generate the coordinates for the next timestep. We do this for each vehicle till the time that the vehicle exits the intersection. This is described in the Figure \ref{fig:data_flow} as "simulation-in-the-loop" method. We run the model in inference mode for 40000 timesteps (4000 seconds), where 1026 vehicles enter the scene. Using "simulation-in-the-loop", SUMO adds multiple vehicles at every timestep and the model must unroll multiple trajectories simultaneously based on the signal timing.

\subsection{Results}
We have provided results on the above metrics only for East-Bound(EB), West-Bound(WB) and North-Bound(NB) lanes as shown in Figure \ref{fig:map} as the South-Bound road is a one-way street.  We run the inference on our current model, which uses multi-head attention to dynamically react to different agents surrounding it. We tabulate the metric results in Table \ref{tab:result_new_model}. We then compare these results with the metrics we have calculated for the model we published in \cite{Ranjan2025PAKDD}, which incorporates an intersection-based vehicle position encoder as well as traffic signal state. We tabulate the metric results in Table \ref{tab:result_old_model}. We also tabulate the results on running the metric on ground truth trajectories generated by SUMO simulator in Table \ref{tab:result_sumo}.
The results of our new model considerably out-performs the results of the earlier model on Mid-Intersection Stoppage (19.2\% decrease), Unsafe Deceleration (9.1\% decrease), Reversing (5.6\% decrease) and TTC events (37\% decrease) metrics. These results are comparable to the ground truth metrics results on Mid-Intersection, Pre-Stopbar Stoppage, Unsafe Deceleration, and Reversing metrics.

\section{Conclusion and Discussion}
In this paper, we have proposed a framework to incorporate intersection-centric actors like signal timing, infrastructure geometry and neighboring vehicles into a trajectory prediction model, as well as evaluate the model by running it in "simulation-in-loop", on traffic-related metrics. We have also provided a multi-headed attention-based trajectory prediction model that uses vector representation to encode the intersection architecture and neighboring actors and dynamically assigns weights to it. The current model considerably outperforms the older model on Mid-Intersection Stoppage, Pre-Stopbar Stoppage and Unsafe De-acceleration. This is because of the efficient representation of lane centerline and signal stopline using a vectorized representation that constrains the distribution of the predicted trajectories to the drivable region and reduces the number of infeasible trajectories. There is a significant reduction in the TTC events, which are the near misses. This is because the multi-headed attention mechanism learns to assign importance to nearby vehicles (which are in front of it) so that the vehicle starts decelerating at the right distance in order to maintain a safe distance.

The model struggles to learn to stop at the red light for west-bound traffic. We think that even though the attention-based encoder is able to learn a good representation of the data, the generated latent mixture and distribution by the decoder are not able to capture this modality of vehicles. A future direction of research could be to try out a transformer-based decoder that uses cross-attention to capture the correlation of the future states with the dynamic and static actors. It would also be interesting to compare the performance of temporal attention with LSTM to capture the temporal correlation between the state of the vehicle.

%
%
%
\bibliographystyle{splncs04}
 \bibliography{itsc2025}

\begin{thebibliography}{10}
\providecommand{\url}[1]{\texttt{#1}}
\providecommand{\urlprefix}{URL }
\providecommand{\doi}[1]{https://doi.org/#1}

\bibitem{7780479}
Alahi, A., Goel, K., Ramanathan, V., Robicquet, A., Fei-Fei, L., Savarese, S.: Social lstm: Human trajectory prediction in crowded spaces. In: 2016 IEEE Conference on Computer Vision and Pattern Recognition (CVPR). pp. 961--971 (2016). \doi{10.1109/CVPR.2016.110}

\bibitem{370fbeadb5584ba9ab2938431fc4f140}
Bishop, C.: Mixture density networks. Workingpaper, Aston University (1994)

\bibitem{chen2023datadriventrafficsimulationcomprehensive}
Chen, D., Zhu, M., Yang, H., Wang, X., Wang, Y.: Data-driven traffic simulation: A comprehensive review (2023), \url{https://arxiv.org/abs/2310.15975}

\bibitem{Fellendorf2010}
Fellendorf, M., Vortisch, P.: Microscopic Traffic Flow Simulator VISSIM, pp. 63--93. Springer New York, New York, NY (2010). \doi{10.1007/978-1-4419-6142-6\_2}, \url{https://doi.org/10.1007/978-1-4419-6142-6\_2}

\bibitem{gao2020vectornetencodinghdmaps}
Gao, J., Sun, C., Zhao, H., Shen, Y., Anguelov, D., Li, C., Schmid, C.: Vectornet: Encoding hd maps and agent dynamics from vectorized representation (2020), \url{https://arxiv.org/abs/2005.04259}

\bibitem{graves2014generatingsequencesrecurrentneural}
Graves, A.: Generating sequences with recurrent neural networks (2014), \url{https://arxiv.org/abs/1308.0850}

\bibitem{gu2021densetntendtoendtrajectoryprediction}
Gu, J., Sun, C., Zhao, H.: Densetnt: End-to-end trajectory prediction from dense goal sets (2021), \url{https://arxiv.org/abs/2108.09640}

\bibitem{ivanovic2019}
Ivanovic, B., Pavone, M.: The trajectron: Probabilistic multi-agent trajectory modeling with dynamic spatiotemporal graphs (2019), \url{https://arxiv.org/abs/1810.05993}

\bibitem{kingma2022autoencodingvariationalbayes}
Kingma, D.P., Welling, M.: Auto-encoding variational bayes (2022), \url{https://arxiv.org/abs/1312.6114}

\bibitem{etde_627062}
Krauss, S.: Microscopic modeling of traffic flow: investigation of collision free vehicle dynamics (Apr 1998)

\bibitem{9234702}
Lefkopoulos, V., Menner, M., Domahidi, A., Zeilinger, M.N.: Interaction-aware motion prediction for autonomous driving: A multiple model kalman filtering scheme. IEEE Robotics and Automation Letters  \textbf{6}(1),  80--87 (2021). \doi{10.1109/LRA.2020.3032079}

\bibitem{SUMO2018}
Lopez, P.A., Behrisch, M., Bieker-Walz, L., Erdmann, J., Fl{\"o}tter{\"o}d, Y.P., Hilbrich, R., L{\"u}cken, L., Rummel, J., Wagner, P., Wie{\ss}ner, E.: Microscopic traffic simulation using sumo. In: The 21st IEEE International Conference on Intelligent Transportation Systems. IEEE (2018), \url{https://elib.dlr.de/124092/}

\bibitem{oshea2015introductionconvolutionalneuralnetworks}
O'Shea, K., Nash, R.: An introduction to convolutional neural networks (2015), \url{https://arxiv.org/abs/1511.08458}

\bibitem{Ranjan2025PAKDD}
Ranjan, Y., Sengupta, S., Rangarajan, R., Ranka, S.: Evaluating generative vehicle trajectory models for traffic intersection dynamics (2025), accepted for publication at a Special Session of PAKDD 2025

\bibitem{salzmann2021}
Salzmann, T., Ivanovic, B., Chakravarty, P., Pavone, M.: Trajectron++: Dynamically-feasible trajectory forecasting with heterogeneous data (2021), \url{https://arxiv.org/abs/2001.03093}

\bibitem{vehits23}
Sengupta, R., Banerjee, T., Karnati, Y., Ranka, S., Rangarajan, A.: Using dsrc road-side unit data to derive braking behavior. In: Proceedings of the 9th International Conference on Vehicle Technology and Intelligent Transport Systems - VEHITS. pp. 420--427. INSTICC, SciTePress (2023). \doi{10.5220/0012025300003479}

\bibitem{shi2023motiontransformerglobalintention}
Shi, S., Jiang, L., Dai, D., Schiele, B.: Motion transformer with global intention localization and local movement refinement (2023), \url{https://arxiv.org/abs/2209.13508}

\bibitem{vaswani2023attentionneed}
Vaswani, A., Shazeer, N., Parmar, N., Uszkoreit, J., Jones, L., Gomez, A.N., Kaiser, L., Polosukhin, I.: Attention is all you need (2023), \url{https://arxiv.org/abs/1706.03762}

\bibitem{viroli2017deepgaussianmixturemodels}
Viroli, C., McLachlan, G.J.: Deep gaussian mixture models (2017), \url{https://arxiv.org/abs/1711.06929}

\bibitem{8186191}
Xie, G., Gao, H., Qian, L., Huang, B., Li, K., Wang, J.: Vehicle trajectory prediction by integrating physics- and maneuver-based approaches using interactive multiple models. IEEE Transactions on Industrial Electronics  \textbf{65}(7),  5999--6008 (2018). \doi{10.1109/TIE.2017.2782236}

\bibitem{8569595}
Xin, L., Wang, P., Chan, C.Y., Chen, J., Li, S.E., Cheng, B.: Intention-aware long horizon trajectory prediction of surrounding vehicles using dual lstm networks. In: 2018 21st International Conference on Intelligent Transportation Systems (ITSC). pp. 1441--1446 (2018). \doi{10.1109/ITSC.2018.8569595}

\bibitem{zhao2020tnttargetdriventrajectoryprediction}
Zhao, H., Gao, J., Lan, T., Sun, C., Sapp, B., Varadarajan, B., Shen, Y., Shen, Y., Chai, Y., Schmid, C., Li, C., Anguelov, D.: Tnt: Target-driven trajectory prediction (2020), \url{https://arxiv.org/abs/2008.08294}

\bibitem{zhao2018infovaeinformationmaximizingvariational}
Zhao, S., Song, J., Ermon, S.: Infovae: Information maximizing variational autoencoders (2018), \url{https://arxiv.org/abs/1706.02262}

\end{thebibliography}

\end{document}